\documentclass{article}

\usepackage[preprint]{neurips_2019}

\usepackage[utf8]{inputenc} 
\usepackage[T1]{fontenc}    
\usepackage{hyperref}  
\usepackage{booktabs}       
\usepackage{amsfonts}       
\usepackage{nicefrac}       
\usepackage{microtype}      
\usepackage{amsmath}
\usepackage{tikz}
\usetikzlibrary{pgfplots.groupplots}
\usepackage{pgfplots}
\usepackage{textcomp}
\usepackage{subfig}
\usepackage{caption}
\usepackage{tikzscale}
\usepackage{appendix}
\usepackage{numprint}
\usepackage{tabularx}
\usepackage{verbatim}
\usetikzlibrary{arrows}
\usetikzlibrary{shapes.geometric}
\usetikzlibrary{matrix,calc,fit}
\usepackage{booktabs}
\usepackage{footnote}

\newcommand\eps{\varepsilon}

\def\GG{E}
 \def\xx{\mathbf{x}}

\pgfplotsset{compat=1.14}
\begin{document}
\title{Generative Latent Flow}

%
\author{Zhisheng Xiao\thanks{equal contribution}, Qing Yan \footnotemark[1]\\
University of Chicago\\
Chicago, IL 60637, USA \\
\texttt{\{zxiao, yanq\}@uchicago.edu} \\
\And
Yali Amit \\
University of Chicago\\
Chicago, IL 60637, USA \\
\texttt{amit@marx.uchicago.edu} \\
}

\maketitle


\begin{abstract}
  In this work, we propose the Generative Latent Flow (GLF), an algorithm for generative modeling of the data distribution. GLF uses an Auto-encoder (AE) to learn latent representations of the data, and a normalizing flow to map the distribution of the latent variables to that of simple i.i.d noise. In contrast to some other Auto-encoder based generative models, which use various regularizers that encourage the encoded latent distribution to match the prior distribution, our model explicitly constructs a mapping between these two distributions, leading to better density matching while avoiding over regularizing the latent variables. We compare our model with several related techniques, and show that it has many relative advantages including fast convergence, single stage training and minimal reconstruction trade-off. We also study the relationship between our model and its stochastic counterpart, and show that our model can be viewed as a vanishing noise limit of VAEs with flow prior.  Quantitatively, under standardized evaluations, our method achieves state-of-the-art sample quality among AE based models on commonly used datasets, and is competitive with GANs' benchmarks. 
\end{abstract}

\section{Introduction} \label{intro}
Deep generative models have recently attracted much attention in the deep learning literature. These models are used to formulate the distribution of complex data as a function of random noise passed through a network, so that rendering samples from the distribution is particularly easy. Deep generative models can be roughly classified into explicit and
implicit models. The former class assumes explicit parametric specification of the distribution, whereas the latter does not. Implicit models are dominated by Generative Adversarial Networks (GANs) \citep{GAN,DCGAN}. GANs have exhibited impressive performance in generating high quality images \citep{biggan} and other vision tasks \citep{cycgan,super}. Despite their successes, training GANs can be challenging, partly because they are trained by solving a saddle point optimization problem formulated as an adversarial game. It is well known that training GANs is unstable and extremely sensitive to hyper-parameter settings \citep{improve,equi}, and sometimes training leads to mode collapse \citep{tutorial}, where most of the samples share some common properties. Although there have been multiple efforts to overcome the difficulties in training GANs, by modifying the objective functions or introducing normalization \citep{wgan,unroll,VEE,spectralnorm}, researchers are also actively studying non-adversarial methods that are known to be less affected by these issues.  

Some explicit methods directly model $p(x)$, the distribution of data, and training is guided by maximizing the data likelihood. For example, auto-regressive models \citep{MAF,PIXELRNN}, which assume the data distribution can be expressed in an auto-regressive pattern, have a simple and stable training process, and currently give the best likelihood results; however, they cannot provide low-dimensional representations of images, and their sampling procedure is inefficient. Normalizing flows \citep{NICE,REAL,GLOW} model $p(x)$ as an invertible transformation from a simple distribution through a change of variables. While being mathematically clear, normalizing flows have one major drawback: computational complexity. Flows have to keep the dimensionality of the original data in order to maintain bijectivity, and this makes training computationally expensive. Considering the prohibitively long training time and advanced hardware requirements in training large scale flow models such as \citep{GLOW}, we believe that it is worth exploring the application of flows in the low dimensional representation spaces rather than in the original data.  

Other explicit generative models often adopt low dimensional latent representations, which are usually obtained from auto-encoders, and generate samples by decoding $z$'s sampled from a pre-defined prior distribution $p(z)$. We call this type of method AE based models. Variational Auto-encoders (VAEs) \citep{VAE,rezende} are perhaps the most influential AE based models. VAEs are trained to minimize a variational bound of the data log likelihood, which is composed of the reconstruction loss plus the KL divergence between $q(z|x)$, the approximate posterior distribution returned by the probabilistic encoder, and the prior $p(z)$. AE based models are easy to train, and they provide low dimensional codes for the data, but unfortunately, their generation quality still lies far below that of GANs, especially on large datasets. For example, it is observed that VAEs tend to generate blurry images, an effect that is usually attributed to the failure to match the marginal distribution in the latent space \citep{TwoVAE,match}. Some modifications to VAEs \citep{IWAE,impFlow1} improve the estimated test data likelihood. However, it is known that higher likelihood is not directly related to better sample quality \citep{note,flowgan}. Some other modifications have recently been proposed  to better match the distribution of latent variables and the prior distribution \citep{TwoVAE,wae,From, NAIS}, and they are shown to have the potential to generate high quality samples. These are discussed in greater detail in Section \ref{rw}. 

Our work pursues the same goal of improving the generation quality of AE based models. To this end, we propose Generative Latent Flow (GLF), which uses a deterministic auto-encoder to learn a mapping to and from a latent space, and a normalizing flow that serves as an invertible transformation between the latent space distribution and a simple noise distribution. Our contributions are summarized as follows: i) we propose Generative Latent Flow, which is an AE based generative model that can generate high quality samples. ii) through standardized evaluations, we show that our model achieves state-of-the-art sample quality among competing models, and can match the benchmarks of GANs. Moreover it has the advantage of one stage training and faster convergence. iii) we carefully study some variants of our method and show its relationship to other methods.

\section{Motivation and related works} \label{rw}
Consider an AE based generative model that can generate samples from the data space $\mathcal{X}$. Ideally, the auto-encoder defines a low dimensional latent space $\mathcal{Z}$, where each image $x_i \in \mathcal{X}$ is associated with a latent vector $z_i \in \mathcal{Z}$ through the decoder $x_i = G(z_i)$. The marginal distribution over $\mathcal{Z}$, denoted by $\tilde p(z)$, is unknown and possibly complicated, and depends on the encoder $E$. In some cases, the model also has a predefined prior distribution $p(z)$ on $\mathcal{Z}$. Since samples are generated by sampling $\mathbf{\eps} \sim p(\eps)$ from the prior and feeding $\mathbf{\eps}$ to the decoder,  in order to generate high quality samples, AE based models need to have: (a) a good decoder $G$ that can output realistic images given latent variables sampled from $\tilde p(z)$, and (b) a good match between $\tilde p(z)$ and $p(z)$. 

Criterion (a) is easily ensured by minimizing the reconstruction loss of the auto-encoders, and there are different ways to ensure criterion (b). Intuitively, criterion (b) can be achieved by either modifying the encoder so that $\tilde p(z)$ is close to $p(z)$, or conversely modifying $p(z)$ to match some observed distribution on the latent space. The classic VAE model adopts the first approach indirectly using an approximation $q(z)$ for $\tilde p(z)$. It assumes a simple prior, and the KL regularizer in the ELBO objective penalizes $KL\left[q(z) \| p(z)\right]$ plus a mutual information term as shown in \citep{surgery}. The approximation $q(z) = \mathbb{E}_{x \sim p_{data}}\left[q(z|x)\right]$ is called the aggregated approximate posterior. Several modifications to VAE's objective \citep{surgery,tcvae, factorsing}, which are designed for the task of unsupervised disentanglement, further decompose the KL term in ELBO, put a stronger penalty specifically on the mismatch between $q(z)$ and $p(z)$. There are also attempts to use normalizing flows as VAEs' posterior distributions \citep{impFlow1,impFlow2,impFlow3}. Although similar in name, they are completely different from our models, as they aim to complicate the approximate posterior $q(z|x)$ thus modifying the distribution $q(z)$. As of yet, these modifications to VAEs have not been shown to improve generation quality. In particular, empirically VAEs with flow posterior have been shown to improve neither the matching of $q(z)$ and $p(z)$ \citep{match}, nor the generation quality \citep{TwoVAE}. Adversarial auto-encoders \citep{AAE} and Wasserstein auto-encoders \citep{wae} use adversarial regularizer or MMD regularizers \citep{MMD} to force the $q(z)$ to be close to $p(z)$. These regularizations can be applied to both deterministic and probabilistic auto-encoders, and are shown to improve generation quality, as they generate sharper images than VAEs do.  

One problem with regularizing $q(z)$ is that it introduces a trade off with reconstruction, i.e. criterion (a). This motivates the use of learnable priors optimized to match $q(z)$. \citep{vamp, hiera,resample} propose different ways to approximate the aggregated posterior $q(z)$ during training, and use the approximated $q(z)$ as their VAEs' prior distributions. This is a natural way to modify the prior to match $q(z)$, however, these methods have not been shown to improve generation quality. Two-stage-VAE\citep{TwoVAE} introduces another VAE on the latent space defined by the first VAE to learn the distribution of its latent variables. GLANN \citep{NAIS} learns a latent representation by GLO \citep{GLO} and matches the densities of the latent variables with an implicit maximum likelihood estimator \citep{IMLE}. VQ-VAE \citep{VQVAE} first learns an AE with discrete latent variables that are stored in a code-book, and then fits an auto-regressive prior on the latent space. RAE+GMM\citep{From} trains a regularized auto-encoder \citep{RAE} and fits a mixture of Gaussian distribution on the latent space. These methods significantly improve the generation quality, but they all involve two-stage training procedure, which adds a computational overhead.

Motivated by above works, we wish to design an AE based generative model that enjoys the best of both worlds: it can be trained end-to-end in a single stage, and it can greatly improve the generation quality without over regularizing the latent variables. We accomplish these goals by using normalizing flow on the latent space of a deterministic AE. More details will be presented in Section \ref{model}. Note that our method is closely related VAEs with normalizing flow as a learnable prior and the connections are discussed in Section \ref{connect}.

\begin{figure}[ht]
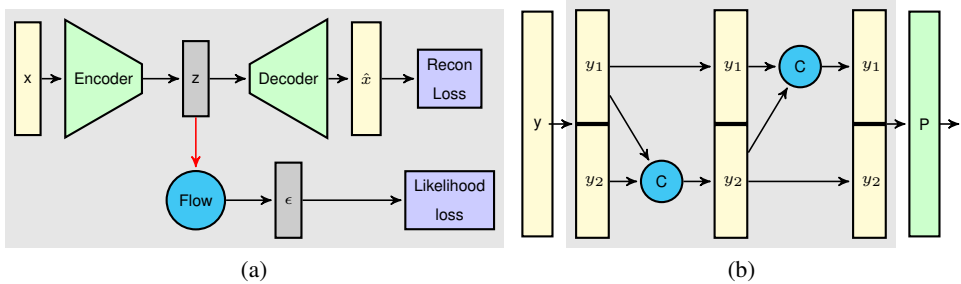

\centering
\subfloat[\label{fig:1a}]{\includegraphics{plot_tikz/g1}}\hspace{4pt}
\subfloat[\label{fig:1b}]{\includegraphics{plot_tikz/g2}}
\caption{(a) Illustration of GLF model. The red arraw contains a stop gradient operation. See Section \ref{strategies}. (b) Structure of one flow block. It splits the input into two parts $y = (y_1,y_2)$, goes through the coupling layer $C$, and applies the random permutation $P$.} \label{fig:1}
\end{figure}

\section{The Generative latent flow (GLF) model} \label{model}
Our model uses a deterministic auto-encoder composed of an encoder $\GG_\eta:\mathcal{X}\rightarrow \mathcal{Z}$ and decoder (generator) $G_\phi: \mathcal{Z} \rightarrow \mathcal{X}$. In addition we have a transformation $F_\theta$  from the distribution on the noise space $\mathcal{E}$, which is assumed to be the standard Gaussian distribution, to the distribution on $\mathcal{Z}$. The transformation $F_\theta$ is defined in terms of a normalizing flow and all three components are learned simultaneously end to end using a loss that combines the reconstruction quality and the likelihood of the encoded data $z_i$ with respect to the transformation $F_\theta$.
 
\subsection{Normalizing flows for the transformation \texorpdfstring{$F$}{F}}
 The core of normalizing flows is carefully-designed invertible networks that map the training data to a simple distribution. Let $z \in \mathcal{Z}$ be an observation from an unknown target distribution $z\sim p(z)$ and $p_{\eps}$ be the unit Gaussian prior distribution on $\mathcal{E}$. Given a bijection $f_{\theta}:\mathcal{Z}\rightarrow\mathcal{E}$, we define a model $p_{\theta}(z)$ with parameters $\theta$ on $\mathcal{Z}$, and we can compute the negative log likelihood (NLL) of $z$ by the change of variable formula: 
 \begin{align}\label{flow}
     -\log(p_{\theta}(z)) = \mathcal{L}_{\text{NLL}}(f_{\theta}(z))=-\Big(\log{p_{\eps}(f_{\theta}(z))}+\log{\left|{\rm{det}}\left(\frac{\partial f_{\theta}(z)}{\partial z}\right)\right|} \Big)
\end{align}
where $\frac{\partial f_{\theta}(z)}{\partial z}$ is the Jacobian matrix of $f_{\theta}$. In order to learn the flow $f_{\theta}$, the NLL objective of $z$ is minimized, which is equivalent to maximize the likelihood of $z$. Since the mapping is a bijection, sampling from the trained model $p_{\theta}(z)$ is trivial: simpy sample $\eps \sim p_{\eps}$ and compute $z=f_{\theta}^{-1}(\eps)$. In our method, we use the normalizing flow to model the transformation $F_\theta=f_\theta^{-1}$.

The key to designing a tractable flow model is defining the transformation $f_{\theta}$ so that the inverse transformation and the determinant of the Jacobian matrix can be efficiently computed. Based on \citep{REAL}, we adopt the following layers to form the flows used in our model.

{\bf Affine coupling layer:}
Given a $D$ dimensional input data $z$ and $d<D$, we partition the input into two vectors $z_{1}=z_{1:d}$ and $z_{2}=z_{d+1:D}$. The output of one affine coupling layer is given by
$y_{1}=z_{1}$, $y_{2}=z_{2}\odot{\rm{exp}}(s(z_{1}))+t(z_{1})$ where $s$ and $t$ are functions from $\mathbb{R}^{d}\rightarrow\mathbb{R}^{D-d}$ and $\odot$ is the element-wise product. The inverse of the transformation is explicitly given by
$z_{1}=y_{1}$, $z_{2}=(y_{2}-t(y_{1}))\odot{\rm{exp}}(-s(y_{1}))$. The determinant of the Jacobian matrix of this transformation is simply  $\rm{det}\frac{\partial y}{\partial z}=\prod_{j=1}^d({\rm{exp}}[s(z_{1})_j])$. Since computing both the inverse and the Jacobian does not require computing the inverse and Jacobian of $s$ and $t$, both functions can be arbitrarily complex. 

{\bf Combining coupling layers with random permutation:}
Affine coupling layers leave some components of the input data unchanged. In order to transform all the components, two coupling layers are combined in an alternating pattern to form a coupling block, so the unchanged components in the first layer can be transformed in the second layer. In particular, we add a fixed random permutation of dimensions of the input data after each coupling layer. See Figure \ref{fig:1b} for an illustration of a coupling block used in our model. 

 \subsection{The objective function} \label{objective}
 Having defined the invertible flow $F_\theta: \mathcal{E} \rightarrow \mathcal{Z}$, we also need to train a deterministic auto-encoder composed of an encoder $\GG_\eta$ and a decoder $G_\phi$. The auto-encoder is trained to minimize the reconstruction loss, which we set to be the common MSE loss. The overall training objective is a combination of the reconstruction loss and the NLL loss for the flow transformation:
\begin{align}\label{glf_loss}
	  \mathcal{L}(\eta,\phi,\theta)=  \frac{1}{N}\sum_{i=1}^{N}\Big(\beta\mathcal{L}_{\text {recon}}\big(\xx_i,G_\phi(\GG_\eta (\xx_i))\big) + \mathcal{L}_{\text{NLL}}\big(f_{\theta}(\text{sg}\left[\GG_\eta(\xx_i)\right]) \big)\Big)
\end{align}
where $\eta,\phi$ are the parameters of the encoder and decoder respectively, $\theta$ is the parameter of the flow, $\text{sg}[\cdot]$ is the stop gradient operation, and $\beta$ is a hyper-parameter that controls the relative weight of the reconstruction loss and the NLL loss in equation \eqref{flow}. Note that when assuming the decoder distribution is Gaussian with variance $\gamma^2$ at each pixel, $\beta$ can be viewed as the inverse variance.
 
After training the model, the generating process is easy: first sample a noise $\epsilon \sim \mathcal{N}(0,I)$ and then obtain a generated sample $\Tilde{x} = G_\phi(F(\epsilon))$, where $F=f_\theta^{-1}$. Since the highlight of our model is applying a flow on latent variables, we name it \textbf{Generative Latent Flow} (GLF). See Figure \ref{fig:1a} for an illustration of the GLF model.
 
\subsection{Why Stop The Gradients?}\label{strategies}
The stop gradient operation in \eqref{glf_loss} is important. If we let gradients of the NLL loss back propagate into the latent variables, it can lead to degenerate $z$'s. This is because $f_\theta$ has to transform the $z$'s to unit Gaussian noise, so the smaller the scale of the $z$'s, the more negative the log-determinant of the Jacobian becomes. Since there is no constraint on the scale of latent variables, the Jacobian term can dominate the entire objective, driving the NLL loss to negative infinity through shrinking $z$ towards $0$. While the latent variables cannot become exactly $0$ because of the presence of reconstruction loss in the objective, the extremely small scale of $z$ may cause numerical issues that cause severe fluctuations. Therefore, we propose to stop the gradient of the NLL loss at the latent variables so that it cannot modify the values of $z$ or affect the parameters of the encoder. We experimentally verify the problems of latent regularization in Section \ref{comparison}.

We call our original model with  stopped gradients \textbf{GLF} and without stopped gradients \textbf{regularized GLF}, since the flow acts as a regularizer on the auto-encoder. Note that for  GLF, the value of $\beta$ in \eqref{glf_loss} does not matter, since the reconstruction loss and the NLL loss are independent. Note also that GLF can also be trained two stages, namely an auto-encoder is trained first, and then the flow is trained to map the distributions. Empirically, we find that the two-stage training strategy leads to similar performance, so we only focus on one-stage training. 

\subsection{Connection to VAEs with Flow Prior} \label{connect}
Our method is closely related to VAEs with normalizing flow priors. To see this, consider the ELBO objective of plain VAEs with Gaussian prior and posterior ($\boldsymbol{\eta},\boldsymbol{\phi}$ denote the parameters of encoder and decoder, respectively):
\begin{align} \label{elbo}
\text{ELBO}(\eta, \phi)=\mathbb{E}_{p_{data}(\mathbf{x})} \mathbb{E}_{q_{\eta}(\mathbf{z} | \mathbf{x})}\left[\log p_{\phi,\beta}(\mathbf{x} | \mathbf{z})+\log p(\mathbf{z})-\log q_{\eta}(\mathbf{z} | \mathbf{x})\right]
\end{align}
The first term is related to the reconstruction loss and depends on the precision $\beta$ of the observationd at each pixel, while the last two terms can be combined as $KL\left[q(z | x) \| p(z)\right]$. 

If we introduce a normalizing flow $f_{\theta}$ for the prior distribution, then the prior $p_{\theta}$ becomes $p_{\theta}(\mathbf{z})=p_{\eps}(f_{\theta}(\mathbf{z}))\left|\operatorname{det}\left(\frac{\partial f_{\theta}(\mathbf{z})}{\partial \mathbf{z}}\right)\right|$, where $p_{\eps}$ is the standard Gaussian density. Substituting this prior into \eqref{elbo}, we obtain the $\text{ELBO}(\eta, \phi,\theta)$ for VAEs with flow prior:
\begin{align} \label{elbo_flow}
\mathbb{E}_{p_{data}(\mathbf{x})} \mathbb{E}_{q_{\eta}(\mathbf{z} | \mathbf{x})}\left[\log p_{\phi,\beta}(\mathbf{x} | \mathbf{z})+\log p_{\eps}\left(f_{\theta}(\mathbf{z})\right)+\log \left|\operatorname{det}\left(\frac{\partial f_{\theta}(\mathbf{z})}{\partial \mathbf{z}}\right)\right|-\log q_{\eta}(\mathbf{z} | \mathbf{x})\right]
\end{align}

Comparing \eqref{elbo_flow} and \eqref{glf_loss}, we observe that if the expectation over $q_{\eta}(z|x)$ is estimated by sampling, $\text{ELBO}(\eta, \phi,\theta)$ is precisely the negative of GLF's objective (without stopping gradients) plus an additional entropy term that corresponds to the entropy of encoder distribution. As $\beta$ increases two things occur as demonstrated empirically in Section \ref{comparison}.
First the estimated variances from the encoder decrease, and second the contribution of the reconstruction loss
to the gradient of the encoder parameters becomes larger than the contribution of the flow's likelihood loss.
Thus as $\beta$ increases the VAE+flow converges to GLF. Furthermore Gaussian VAEs with flow prior does not suffer from the degeneracy of regularized GLF
because of the presence of the entropy term. It is the negative sum of the log variances of the latent variables, and thus it encourages the encoder to output large posterior variance, preventing latent variables from collapsing to $0$. 


VAEs with flow prior have attracted very little attention \citep{explicit}, and they have only focused on improvement of the data likelihood. Our work is differs in two ways: 1.  we are the first to evaluate the effects of normalizing flow prior on generation quality; 2. we use deterministic AEs rather than VAEs, thus avoiding the need to choose $\beta$.

\section{Experiments}\label{experiment}
To demonstrate the performance of our method, we present both quantitative and qualitative evaluations on four commonly used datasets for generative models: MNIST \citep{MNIST}, Fashion MNIST \citep{fashion}, CIFAR-10 \citep{CIFAR} and CelebA \citep{CELEB}. Throughout the experiments, we use 20-dimensional latent variables for MNIST and Fashion MNIST, and 64-dimensional latent variables for CIFAR-10 and CelebA. 

\citep{AreGan} adopted a common network architecture based on InfoGAN \citep{InfoGAN} to evaluate GANs. In order to make fair comparisons without designing arbitrarily large networks to achieve better performance, we use the generator architecture of InfoGAN as our decoder's architecture, and we make the encoder to be symmetric to the decoder. For details of the AE network structures, see Appendix \ref{AppA}. For the flow applied on latent variables, we use 4 affine coupling blocks defined as in Figure \ref{fig:1b}, where each block contains 3 fully connected layers each with $k$ hidden units. For MNIST and Fashion MNIST, $k=64$, while for CIFAR-10 and CelebA, $k=256$. Note that the flow only adds a small parameter overhead on the auto-encoder (less than $3\%$).

\subsection{Metrics}
We use the Fr\'echet Inception Distance (FID) \citep{FID} as a metric for image generation quality. FID is computed by first extracting features of a set of real images $x$ and a set of generated images $g$ from an intermediate layer of the Inception network \citep{inception}. Each set of features is fitted with a Gaussian distribution, yielding means $\mu_{x}$, $\mu_{g}$ and co-variances matrices $\Sigma_{x},\Sigma_{g}$. The FID score is defined to be the Fréchet distance between these two Gaussians:
\[\mathrm{FID}(x, g)=\left\|\mu_{x}-\mu_{g}\right\|_{2}^{2}+\operatorname{Tr}\left(\Sigma_{x}+\Sigma_{g}-2\left(\Sigma_{x} \Sigma_{g}\right)^{\frac{1}{2}}\right)\]
It is claimed that the FID score is sensitive to mode collapse and correlates well with human perception of generator quality. Recently, \citep{PRD} proposed using Precision and Recall for Distributions (PRD) which can assess both the quality and diversity of generated samples. We also include PRD in our studies. See Appendix \ref{appD}.

\begin{figure}
    \centering
    \subfloat[MNIST]{\includegraphics[width=0.245\textwidth]{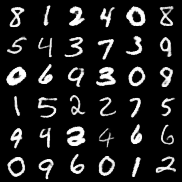}}
    \hfill
    \subfloat[Fashion MNIST]{\includegraphics[width=0.245\textwidth]{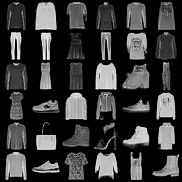}}
    \hfill
    \subfloat[CIFAR-10]{\includegraphics[width=0.245\textwidth]{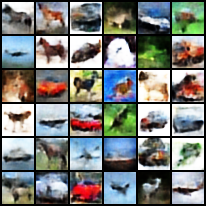}}
     \hfill
    \subfloat[CelebA]{\includegraphics[width=0.245\textwidth]{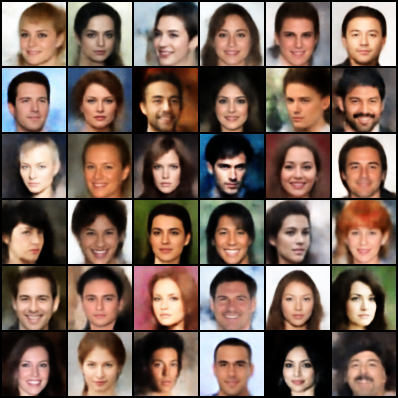}}
 
    \hfill
    \subfloat[CelebA-HQ\label{fig:3e}]{\includegraphics[width=0.44\textwidth]{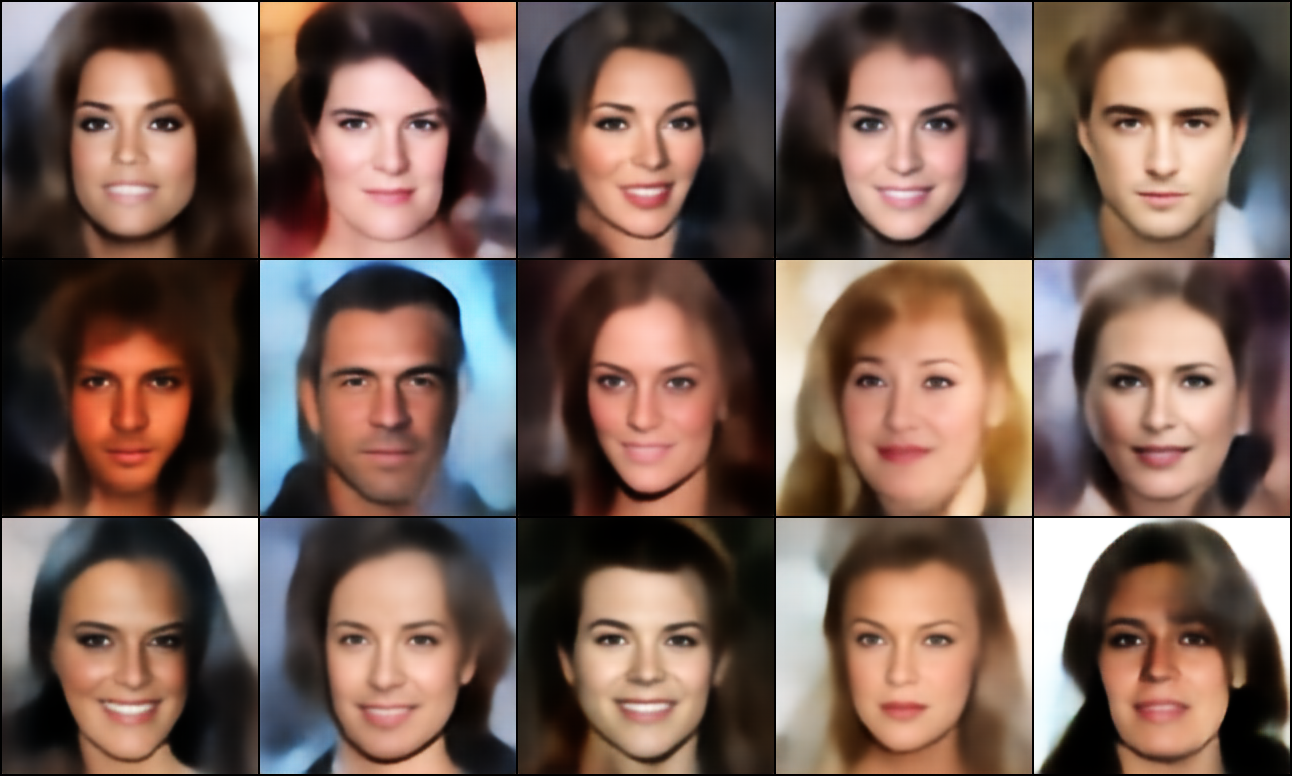}}
     \hfill
    \subfloat[Noise interpolation \label{fig:3f}]{\includegraphics[width=0.48\textwidth]{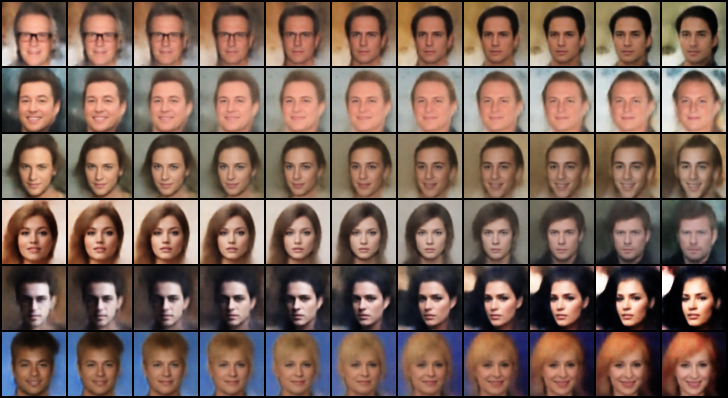}}
    \caption{\label{fig:sample_percept}
    (a)-(e): Randomly generated samples from our method trained on different datasets.
     (f): Random noise interpolation on CelebA.}
\end{figure}

\begin{savenotes}
 \begin{table}[ht]
  \caption{FID scores obtained from different models. For our reported results, we executed 10 independent trials and report the mean and standard deviation of the FID scores. Each trail is computing the FID between 10k generated images and 10k real images.}
  \label{FID}
  \centering
  \begin{tabular}{lllll}
    \toprule
        & MNIST  &Fashion   & CIFAR-10 & CelebA \\
    \midrule
	 VAE & $28.2\pm0.3$ & $57.5\pm0.4$ & $142.5\pm0.6$ & $71.0\pm0.5$\\
	 WAE-GAN  & $12.4\pm0.2$ & $31.5\pm0.4$ & $93.1\pm0.5$ & $66.5\pm0.7$\\
	 Two-Stage VAE  & $10.9\pm0.7$ & $26.1\pm0.9$ & $96.1\pm0.9$\footnote{Note that there is a large discrepancy between this and the result reoprted in the original paper. See Appendix \ref{bug} for explanation} & $65.2 \pm 0.8$ \\
	 RAE + GMM  & $10.8\pm0.1$ & $25.1\pm0.2$ & $91.6\pm0.6$ & $57.8\pm0.4$\\
	 VAE+flow prior & $28.3\pm0.2$ & $51.8\pm0.3$ & $110.4\pm0.5$ & $54.3\pm0.3$\\
	 VAE+flow posterior & $26.7\pm0.3$ & $55.1\pm0.3$ & $143.6\pm0.8$ & $67.9 \pm 0.3$\\
	 GLF (ours) & \textbf{8.2} $\pm$ 0.1 & \textbf{21.3} $\pm$ 0.2 & \textbf{88.3} $\pm$ 0.4 & \textbf{53.2} $\pm$ 0.2 \\
	 \midrule
	 GLANN with perceptual loss & $8.6\pm0.1$ & $13.0\pm0.1$ & $46.5\pm0.2$ & $46.3\pm0.1$ \\
	 GLF+perceptual loss (ours) & \textbf{5.8} $\pm$ 0.1 & \textbf{10.3} $\pm$ 0.1 & \textbf{44.6} $\pm$ 0.3 & \textbf{41.8} $\pm$ 0.2\\
    \bottomrule
  \end{tabular}
\end{table}
\end{savenotes}

\subsection{Results}
Table \ref{FID} summarizes the main results of this work. We compare the FID scores obtained by our method with the scores of the VAE baseline and several existing AE based models that are claimed to produce high quality samples. Instead of directly citing their reported results, we re-ran the experiments because we want to evaluate them under standardized settings so that all models adopt the same AE architectures, latent dimensions and image pre-processing. We use GLF and VAE+flow prior with $\beta=1$ to report the results in the table. For other methods, we largely follow their proposed experimental settings. Details of each experiment are presented in Appendix \ref{AppB}.

Note that the authors of WAE propose two variants, namely WAE-GAN and WAE-MMD. We only report the results of WAE-GAN, as we found it consistently outperforms WAE-MMD. Note also that, GLANN \citep{NAIS} obtains impressive FID scores, but it uses perceptual loss \citep{perceptual} as the reconstruction loss. The perceptual loss is obtained by feeding both training images and reconstructed images into a pre-trained network such as VGG \citep{VGG}, and computing the $L_1$ distance between some of the intermediate layers' activation. We also train our method with perceptual loss and compare with GLANN in the last two rows of Table \ref{FID}.

As shown in Table \ref{FID}, our method obtains significantly lower FID scores than competing AE based models across all four datasets. In particular, GLF greatly outperforms VAE+flow prior in the default setting. A more detailed analysis and comparison between the two methods will be done in Section \ref{comparison}. We also confirm that VAE+flow posterior cannot improve generation quality. Perhaps the competing model with the closest performances to ours is RAE+GMM, which shares some similarity with GLF in that both methods fit the density of the latent variables of an AE explicitly. To compare our method with GANs, we also include the results from \citep{AreGan} in Appendix \ref{GAN compare}. In \citep{AreGan}, the authors conduct standardized and comprehensive evaluations of representative GAN models with large-scale hyper-parameter searches, and therefore, their results can serve as a strong baseline. The results indicate that our method's generation quality is competitive with that of carefully tuned GANs.

Qualitative results are shown in Figure \ref{fig:sample_percept}. Besides samples of the datasets used for quantitative evaluation, samples of CelebA-HQ \citep{celebahq} with the larger size of $256 \times 256$ are also included in Figure \ref{fig:3e} to show our method's potential of scaling up to images with higher resolution. Qualitative results show that our model can generate sharp and diverse samples in each dataset. In Figure \ref{fig:3f}, we show CelebA images generated by linearly interpolating two samples of random noise. The smooth interpolation indicates that our method fits the distribution of latent variables well. For more qualitative results, including samples from the models trained with perceptual loss, see Appendix \ref{AppC}. We see that samples from models trained with perceptual loss have higher quality.

\subsubsection{Comparisons: GLF vs. Regularized GLF and VAE+flow Prior.}\label{comparison}


As discussed in Section \ref{strategies}, regularized GLF is unstable because of the degeneracy of latent variables created by the NLL loss. We empirically study the effect of latent regularization as a function of $\beta$ on CIFAR-10. For low values of $\beta=1$ and $10$, the NLL loss completely dominates the learning signal and the reconstruction loss quickly diverges. Even for larger values of $\beta=50,100,400$
the NLL loss decreases to a very small negative value, and although overall performance is reasonable it oscillates quite strongly as training proceeds. The relevant plots are shown in Figure \hyperref[fig:cifar glf]{4} in Appendix \ref{additional}.
In contrast, for GLF, where the flow does not modify $z$, the NLL loss does not degenerate, resulting in stable improvements of FID scores as training progress. 

\begin{figure}[ht]
\centering
\includegraphics{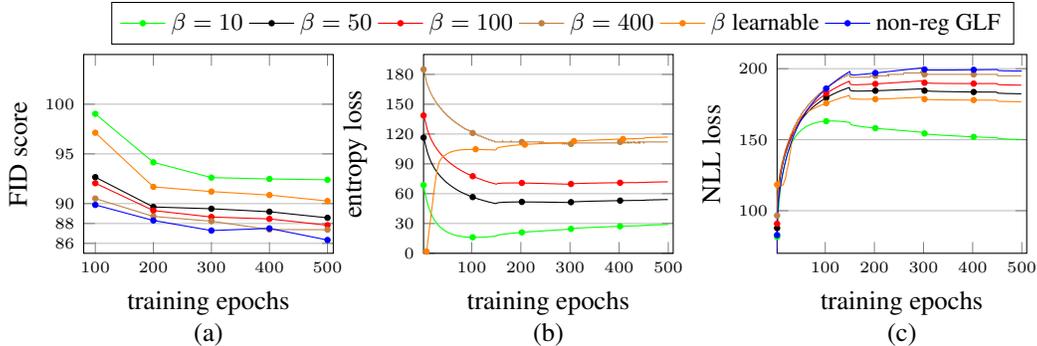}
\caption{(a) Record of FID scores on CIFAR-10 for VAEs+flow prior with different values of $\beta$ and  GLF. (b) Record of entropy losses for corresponding models. (c) Record of NLL losses for corresponding models.} \label{fig:4}. 
\end{figure}

We also trained VAEs+flow prior with different choices of decoder variances (equivalently, different choices of $\beta$), plus one with learnable decoder variance as done in \citep{TwoVAE}. We record the progression of FID scores of these models on CIFAR-10 in Figure \hyperref[fig:4]{3a}.
In Figure \hyperref[fig:4]{3b}, we plot the entropy loss, which is one term in VAE+flow prior's minimization objective. The entropy loss is expressed as $-\sum_{j=1}^{d} \log(\sigma_j)/2$, where $\sigma_j$ is the standard deviation of the $j^{th}$ latent variable. Higher entropy loss means that the latent variables have lower variances. In Figure \hyperref[fig:4]{3c}, we plot the NLL loss. 

From Figure \hyperref[fig:4]{3a}, we see that GLF converges faster and obtains lower FID score than VAEs+flow prior. The performance gap closes as $\beta$ increases, however, even with large $\beta$, GLF still slightly outperforms VAE+flow prior. We also find that the learnable $\beta$ for VAE+flow prior is not effective, probably due to relatively small values of $\beta$ in early time. When $\beta$ is large, as indicated before, the posterior variances become very small, so that effectively we are training an AE. For example, as shown in Figure \hyperref[fig:4]{3b}, when $\beta=400$, the corresponding average posterior variance is around $10^{-4}$. 

In contrast to regularized GLF, there is no degeneracy of latent variables observed thanks to the noise introduced by VAEs and the corresponding entropy term. Indeed, Figure \hyperref[fig:4]{3c} shows that the training of VAE+flow prior does not over-fit the NLL loss, as opposed to regularized GLF where severe over-fitting to NLL loss occurs as shown in Figure \hyperref[fig:cifar glf]{4c}. Comparing Figure \hyperref[fig:4]{3a} and \hyperref[fig:cifar glf]{4a}, we observe that unlike regularized GLF, VAE+flow prior does not suffer from divergence or fluctuations in FID scores, even with relatively small $\beta$. In general, the results of FID scores show that regularized GLF is unstable, while as $\beta$ increases, the performance of VAE+flow prior converges to that of GLF, which outperforms them all.

\subsection{Training time}\label{training}
Besides better performances, our method also has the advantage of faster convergence among competing methods such as GLANN and Two-stage VAE. In Table \ref{epoch} we compare the number of training epochs to obtain the FID scores in Table \ref{FID}. We also compare the per epoch training clock time in Appendix \ref{time}. The results indicate that GLF requires much less training time to generate high quality samples. 

\begin{table}
  \caption{Number of training epochs for Two-stage VAE, GLANN, and GLF}
  \label{epoch}
  \centering
  \begin{tabular}{llll}
    \toprule
        & MNIST/Fashion   & CIFAR-10 & CelebA \\
    \midrule
	 Two-stage VAE First/Second & 400/800 & 1000/2000 &  120/300 \\
	 GLANN First/Second & 500/50  & 500/50 & 500/50 \\
	 GLF & 100 & 200 &  40  \\
    \bottomrule
  \end{tabular}
\end{table}

\section{Conclusion}
In this paper, we introduce Generative Latent Flow, a novel generative model which uses an auto-encoder to learn a latent space from training data and a normalizing flow to match the distribution of the latent variables with the prior. Under standardized evaluations, our model achieves state-of-the-art results in image generation among several recently proposed Auto-encoder based models. Besides higher generation quality, our method also enjoys advantages such as faster training time and end-to-end single stage training. While we are not claiming that our GLF model is superior to GANs, we do believe that it opens the door to realize the potential of AE based models to produce high quality samples just as GANs do. The comparison between our method and its stochastic counterparts briefly examines the question about the effects of adding noise during the training of generative models, which is a topic that deserves further studies.  

\bibliographystyle{plain}

\newpage
\appendix
\appendixpage
\section{Network Architectures} \label{AppA}
In this section we provide Table \ref{netstruct} that summarizes the auto-encoder network structure. The network structure is adopted from InfoGAN\citep{InfoGAN}, and the difference between the networks we used for each dataset is the size of the fully connected layers, which depends on the size of the image. All convolution and deconvolution layers have $\text{stride} = 2$ and $\text{padding}=1$ to ensure the spatial dimension decreases/increases by a factor of 2. $M$ is simply the size of an input image divided by $4$. Specifically, for MNIST and Fashion MNIST, $M=7$; for CIFAR-10, $M=8$; for CelebA, $M=16$. BN stands for batch normalization.

For VAEs, the final FC layer of the encoder will have doubled output size to return both the mean and standard deviation of latent variables. 

\begin{table}
  \caption{Network structure for auto-encoder based on InfoGAN}
  \label{netstruct}
  \centering
  \begin{tabular}{ll}
    \toprule
    Encoder   & Decoder    \\
    \midrule
     Input $x$   &      Input $z$ \\
     $4 \times 4$ $\text{Conv}_{64}$, ReLU   &  FC $\text{nz} \rightarrow 1024$, BN, ReLU\\
     $4 \times 4$ $\text{Conv}_{128}$, BN, ReLU &  FC $1024 \rightarrow 128\times M \times M$, BN, ReLU \\
     Flatten, FC $128\times M \times M \rightarrow 1024$, BN, ReLU & $4 \times 4$ $\text{Deconv}_{64}$, BN, ReLU \\
     FC $1024 \rightarrow \text{nz} $ & $4 \times 4$ $\text{Deconv}_{128}$, Sigmoid\\
    \bottomrule
  \end{tabular}
\end{table}

\section{Experiment Settings} \label{AppB}
In this section, we present the details of our experimental settings for results in Table \ref{FID}. Since the settings for MNIST and Fashion MNIST are the same, we only mention MNIST for simplicity. For GLANN, we directly cite the results from \citep{NAIS}, as their experimental settings is very similar to ours. 

We use the original images in the training sets for MNIST, Fashion MNIST and CIFAR-10. For CelebA, we follow the same pre-processing as in \citep{AreGan}: center crop to $160 \times 160$ and then resize to $64 \times 64$. 

\subsection{Settings for training GLF} \label{AppB1}
For all datasets (except CelebA-HQ), we use batch size 256 and Adam \citep{adam} optimizer with initial learning rate $10^{-3}$ for the parameters of both the AE and the flow. We add a weight decay $2 \times 10^{-5}$ to the optimizers for the flow. For MNIST, we train our model for $100$ epochs, with learning rate decaying by a factor of $2$ after $50$ epochs. For CIFAR-10, we train our model for 200 epochs, with the learning rate decaying by a factor of $2$ every $50$ epochs. For CelebA, we train our model for $40$ epochs with no learning rate decay. 

For GLF with perceptual loss as the reconstruction loss, we compute the perceptual loss as suggested in \citep{NAM}. See \url{https://github.com/facebookresearch/NAM/blob/master/code/perceptual_loss.py} for their implementation. Other settings are the same. 

For CelebA-HQ dataset, we adopt our AE network structure based on DCGAN \citep{DCGAN}. Note that this is a relatively simple network for high resolution imgaes. We use batch size 64, with initial learning rate $10^{-3}$ for both the AE and the flow. We train our model for $60$ epochs, with learning rate decaying by a factor of $2$ after $40$ epochs.

\subsection{Settings for training VAEs and VAE variants}
We adopt common settings for our reported results of VAE, VAE+flow prior and VAE+flow posterior. We still use batch size 256, and Adam optimizer with initial learning rate $10^{-3}$ for both the VAE and the flow, if applicable. We find VAEs need longer time to converge, so we double the training epochs. We train MNIST for $200$ epochs, with learning rate decaying by a factor of $2$ after $100$ epochs. We train CIFAR-10 for $400$ epochs, with the learning rate decaying by a factor of $2$ every $100$ epochs. We train CelebA for $80$ epochs with learning rate decaying by a factor of $2$ after $40$ epochs. 

\subsection{Settings for training WAE-GAN}
We follow the settings introduced in the original WAE paper\citep{wae}. The adversary in WAE-GAN has the following architecture:
\[
\begin{aligned} z \in \mathcal{R}^{d} & \rightarrow \mathrm{FC}_{512} \rightarrow \mathrm{ReLU} \\ & \rightarrow \mathrm{FC}_{512} \rightarrow \operatorname{ReLU} \\ & \rightarrow \mathrm{FC}_{512} \rightarrow \operatorname{ReLU} \\ & \rightarrow \mathrm{FC}_{512} \rightarrow \operatorname{ReLU} \rightarrow \mathrm{FC}_{1} \end{aligned}
\]
where $d$ is the dimension of the latent variables. 

WAE has two major hyper-parameters: $\lambda$ which controls the weight coefficient of the adversarial regularizer, and $\sigma^2$ which is the variance of the prior. Batch size is 100 for all datasets. For MNIST, $\lambda = 10$ and $\sigma^2 = 1$, and the model is trained for 100 epochs. The initial learning rate is $10^{-3}$ for the AE and $5 \times 10^{-4}$ for the adversary. After 30 epochs both learning rates decreased both by factor of 2, and after first 50 epochs further by factor of 5. For CIFAR, $\lambda = 10$ and $\sigma^2 = 1$ and the model is trained for 200 epochs. The initial learning rates are the same as training MNIST, and the learning rate decays by a factor of 2 after first 60 epochs, and further by a factor of 5 after 120 epochs. For CelebA, $\lambda = 1$ and $\sigma^2 = 2$. The model is trained for 55 epochs. The initial learning rate is $3 \times 10^{-4}$ for the AE and $10^{-3}$ for the adversary. Both learning rates decays by factor of 2 after 30 epochs, further by factor of 5 after 50 first epochs.

\subsection{Settings for training Two stage VAE} \label{bug}
We adopt the settings in the original paper \citep{TwoVAE}. For all datasets, the batch size is set to be 100, and the initial learning rate for both the first and the second is $10^{-4}$. For MNIST, the first VAE is
trained for 400 epochs, with learning rate halved every 150 epochs; the second VAE is trained for 800 epochs with learning rate halved every 300 epochs. For CIFAR-10, 1000 and 2000 epochs are trained for the two VAEs respectively, and the learning rates are halved every 300 and 600 epochs for the two stages. For CelebA, 120 and 300 epochs are trained for the two VAEs respectively, and the learning rates are halved every 48 and 120 epochs for the two stages.

\textbf{Explaining the discrepancy between our reported results and the results in the original paper:} The original Two stage VAE paper adopt similar settings with our experiments, but we observe large discrepancies on the results of CIFAR-10 and CelebA. After carefully reviewing their published codes, we find that there is an issue in their FID score computation particularly for CIFAR-10 dataset. Specifically, the true images used for computating the FID on CIFAR-10 is obtained from saving the original data in .jpg format and reading them back, and the saving will cause some errors in pixel values. After fixing this issue, we re-ran their published codes and obtained similar results as we reported. We also run through their testing protocol using samples from our models, and we ontain scores around 65. For CelebA, one particular detail worth noting is that, \citep{TwoVAE} applies $128 \times 128$ center-crop before re-sizing on CelebA, while $160 \times 160$ center-crop is used in our evaluations. With smaller center-crops the human faces occupy a larger portion of the image with less background, making the generative modeling easier.

\subsection{Settings for training RAE+GMM}
The settings of batch size, learning rate scheduling and number of epochs for training RAE are the same as those of GLF. The objective of the RAE is reconstruction loss plus a penalty on the norm of the latent variable. Since the author does not report their choices for the penalty coefficient $\gamma$, we search over $\gamma \in {0.1,0.5,1,2}$, and we find that $\beta=0.5$ leads to the best overall performances, and therefore we let $\gamma = 0.5$. After training the RAE, we fit a 10-component Gaussian mixture distribution on the latent variables.

\section{Comparison with GANs} \label{GAN compare}
In Table \ref{FID_GAN} we combine our reported results of AE based models and the FID scores of GANs cited from \citep{AreGan}.

\begin{table}
  \caption{FID score comparisons of GANs and various AE based models}
  \label{FID_GAN}
  \centering
  \begin{tabular}{lllll}
    \toprule
        & MNIST  &Fashion   & CIFAR-10 & CelebA \\
    \midrule
      MM GAN & $9.8\pm0.9$ & $29.6\pm1.6$ & $72.7\pm3.6$ & $65.6\pm4.2$ \\
	 NS GAN & $6.8\pm0.5$ & $26.5\pm1.6$ & $58.5\pm1.9$ & $55.0\pm3.3$  \\
	 LSGAN & $7.8\pm0.6$ & $30.7\pm2.2$ & $87.1\pm47.5$ & $53.9\pm2.8$  \\
	 WGAN & $6.7\pm0.4$ & $21.5\pm1.6$ & $55.2\pm2.3$ & $41.3\pm2.0$   \\
	 WGAN GP & $20.3\pm5.0$ & $24.5\pm2.1$ & $55.8\pm0.9$ & $30.3\pm1.0$  \\
	 DRAGAN & $7.6\pm0.4$ & $27.7\pm1.2$ & $69.8\pm2.0$ & $42.3\pm3.0$  \\
	 BEGAN & $13.1\pm1.0$ & $22.9\pm0.9$ & $71.4\pm1.6$ & $38.9\pm0.9$  \\
	 \midrule
	 VAE & $28.2\pm0.3$ & $57.5\pm0.4$ & $142.5\pm0.6$ & $71.0\pm0.5$\\
	 WAE-GAN  & $12.4\pm0.2$ & $31.5\pm0.4$ & $93.1\pm0.5$ & $66.5\pm0.7$\\
	 Two-Stage VAE & $10.9\pm0.7$ & $26.1\pm0.9$ & $96.1\pm0.9$ & $65.2 \pm 0.8$ \\
	 RAE + GMM  & $10.8\pm0.1$ & $25.1\pm0.2$ & $91.6\pm0.6$ & $57.8\pm0.4$\\
	 GLANN (with perceptual loss) & $8.6\pm0.1$ & $13.0\pm0.1$ & $46.5\pm0.2$ & $46.3\pm0.1$ \\
	 VAE+flow prior & $28.3\pm0.2$ & $51.8\pm0.3$ & $110.4\pm0.5$ & $54.3\pm0.3$\\
	 VAE+flow posterior & $26.7\pm0.3$ & $55.1\pm0.3$ & $143.6\pm0.8$ & $67.9 \pm 0.3$\\
	 GLF (ours) & $8.2\pm0.1$ & $21.3\pm0.2$ & $88.3\pm0.4$ & $53.2\pm0.2$ \\
	 GLF+perceptual loss (ours) &$5.8\pm0.1$ & $10.3\pm0.1$ & $44.6\pm0.3$ & $41.8\pm0.2$\\
    \bottomrule
  \end{tabular}
\end{table}

\section{Issues with latent regularization}\label{additional}

\begin{figure}[ht]
\centering
\includegraphics{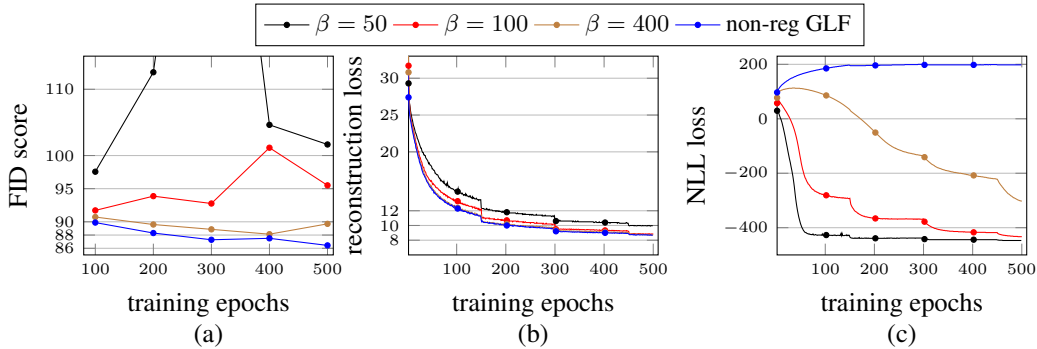}
\caption{(a) Record of FID scores on CIFAR-10 for regularized GLF with different values of $\beta$ and  GLF. $\beta=1$ and $10$ are omitted because they leads to divergence in reconstruction loss. (b) Record of reconstruction losses for corresponding models. (c) Record of NLL losses for corresponding models.}  \label{fig:cifar glf}
\end{figure}

\section{Clock time comparisons}\label{time}
In Table \ref{clock}, we report the clock training time per epoch of our method, two-stage VAE and GLANN. Note that for methods using perceptual loss, the per epoch training time is longer because activations need to be computed. This, together with Table \ref{epoch}, shows that we need much shorter training time while obtaining better performances. In our method, training the flow does not add much computational time due to the low dimensionality.We record the clock time on a platform with a single GTX 1080 GPU. 

\begin{table}[ht]
  \caption{Per-epoch training time in seconds}
  \label{clock}
  \centering
  \begin{tabular}{llll}
    \toprule
        & MNIST & CIFAR-10 & CelebA \\
    \midrule
	 
     2-stage VAE 1st/2nd  & $5/2$ & $6/2$ & $60/28$\\
     GLF & $10$ & $13$ & $108$   \\
     GLANN with perceptual loss& $14$ & $16$ & $292$\\
     GLF with perceptual loss & $16$ & $19$ & $343$ \\
    \bottomrule
  \end{tabular}
\end{table}

\section{Precision and Recall}\label{appD}
In this section, we report the precision and recall (PRD) evaluation of our randomly generated samples on each dataset in Table \ref{prd}. The two numbers in each entry are $F_8, F_{\frac{1}{8}}$ that captures recall and precision. See \citep{PRD} for more details. We report the PRD for our models under the setting of obtaining the results in Table \ref{FID}. 

\begin{table}[ht]
  \caption{ Evaluation of random sample quality by precision / recall. Higher numbers are better. }
  \label{prd}
  \centering
  \begin{tabular}{llll}
    \toprule
         MNIST  &Fashion & CIFAR-10 & CelebA \\
    \midrule
      
	 $(0.981,0.963)$ &  $(0.974,0.955)$ & $(0.685,0.805)$  & $(0.447,0.726)$  \\
    \bottomrule
  \end{tabular}
\end{table}

\section{More qualitative results}\label{AppC}
\begin{figure}[ht]
    \centering
    
%

  \centering
    \subfloat[MNIST]{\includegraphics[width=0.245\textwidth]{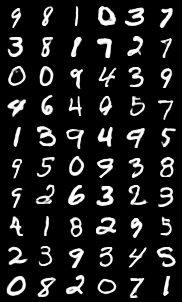}}
    \hfill
    \subfloat[Fashion MNIST]{\includegraphics[width=0.245\textwidth]{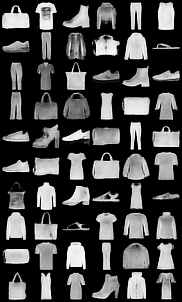}}
    \hfill
    \subfloat[CIFAR-10]{\includegraphics[width=0.245\textwidth]{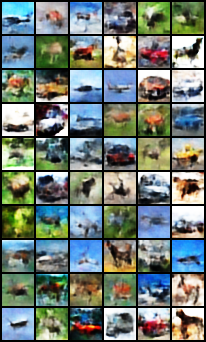}}
     \hfill
    \subfloat[CelebA]{\includegraphics[width=0.245\textwidth]{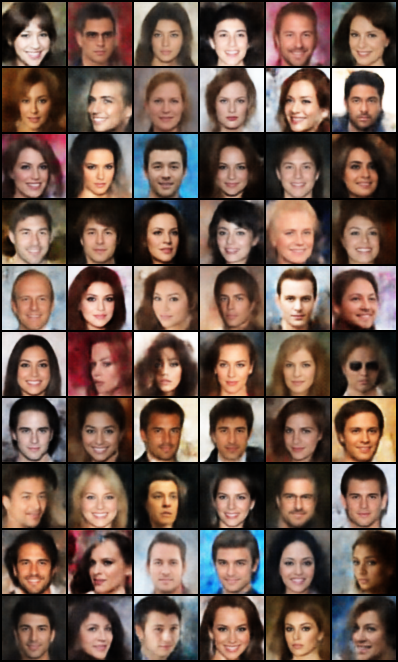}}
    
    \hfill
    
    \subfloat[MNIST]{\includegraphics[width=0.245\textwidth]{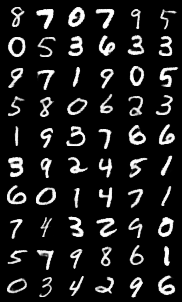}}
    \hfill
    \subfloat[Fashion MNIST]{\includegraphics[width=0.245\textwidth]{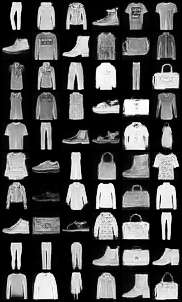}}
    \hfill
    \subfloat[CIFAR-10]{\includegraphics[width=0.245\textwidth]{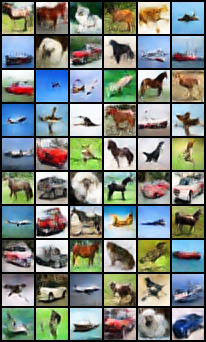}}
     \hfill
    \subfloat[CelebA]{\includegraphics[width=0.245\textwidth]{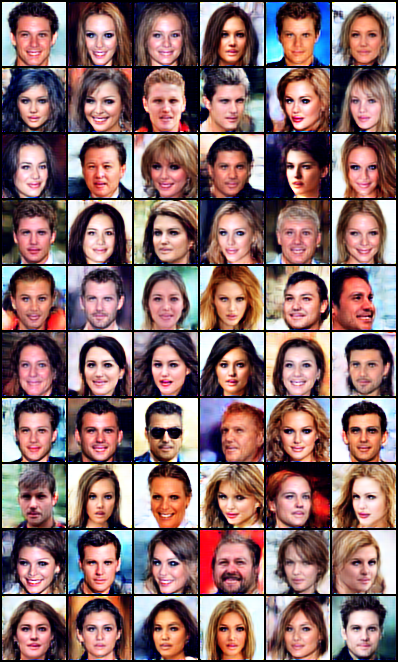}}
    \caption{\label{fig:more_sample_percept1}
    (a)-(d) Randomly generated samples from our method with MSE loss.
    (e)-(h) Randomly generated samples from our method with perceptual loss.}
\end{figure}

\begin{figure}
\centering
\includegraphics[width=0.8\linewidth]{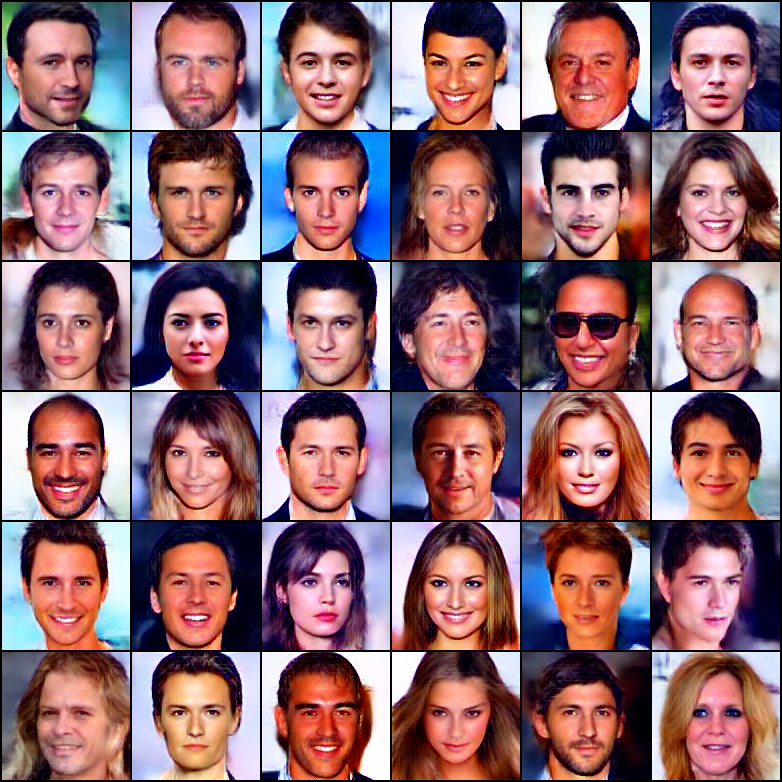}
\caption{Randomly generated samples from our method with perceptual loss on CelebA-HQ dataset} \label{fig:hq2}
\end{figure}

\begin{figure}
\centering
\includegraphics[width=0.8\linewidth]{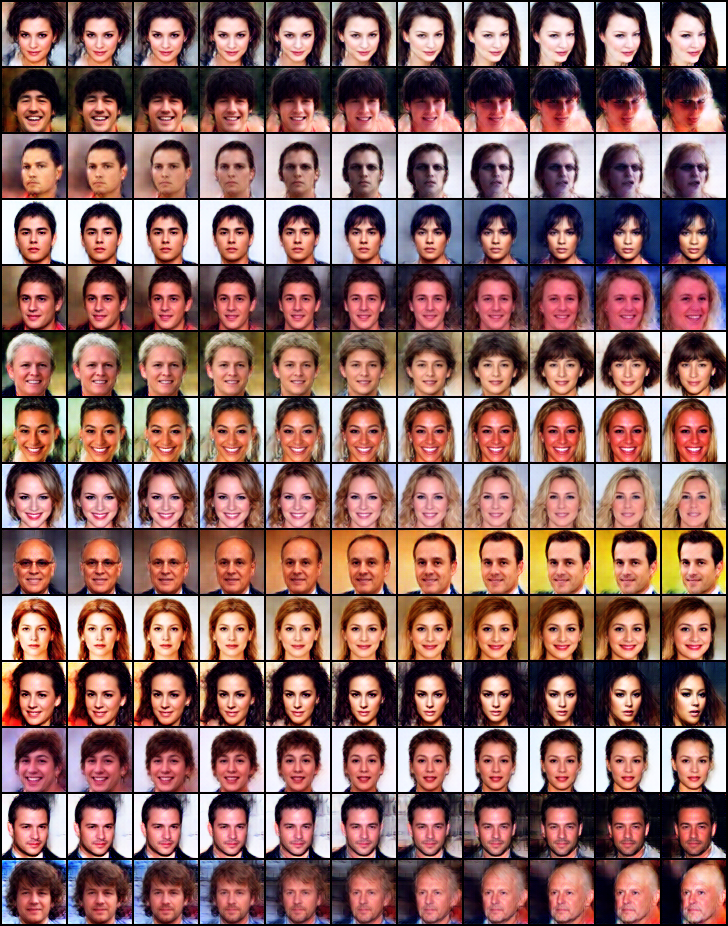}
\caption{Noise interpolation on CelebA} \label{fig:inter}
\end{figure}

In Figure \ref{fig:more_sample_percept1}, we show more samples of each dataset generated by GLF, using either MSE or perceptual loss as reconstruction loss. In Figure \ref{fig:hq2}, we show samples of CelebA-HQ datasets from GLF trained with perceptual loss. In Figure \ref{fig:inter}, we show examples of interpolations between two randomly sampled noises on CelebA from GLF trained with perceptual loss. 

\clearpage

\end{document}